\pdfoutput=1

\documentclass[11pt]{article}

\usepackage[final]{acl}

\usepackage{times}
\usepackage{latexsym}
\usepackage{CJKutf8} 
\usepackage{tcolorbox}
\usepackage{soul}

\usepackage{enumitem}
\usepackage{array}
\usepackage{colortbl}
\usepackage{xcolor}
\usepackage{caption}
\usepackage{tabularray}
\usepackage[utf8]{inputenc} 
\usepackage[T1]{fontenc}    
\usepackage{hyperref}       
\usepackage{url}            
\usepackage{booktabs}       
\usepackage{amsfonts}       
\usepackage{nicefrac}       
\usepackage{microtype}      
\usepackage{graphicx}

\usepackage{natbib}
\usepackage{verbatim}
\usepackage{inconsolata}
\usepackage{tabularx}
\usepackage{float}
\usepackage{algorithmic}
\usepackage{multirow}
\usepackage{listings}
\usepackage{subfigure}
\usepackage{subcaption}
\usepackage{amsmath}
\usepackage{amssymb}
\usepackage{array}
\usepackage[ruled,linesnumbered]{algorithm2e}
\usepackage[T1]{fontenc}
\definecolor{darkgray}{rgb}{0.4, 0.4, 0.4}
\definecolor{backcolour}{rgb}{0.95,0.95,0.92}
\definecolor{myblue}{rgb}{0.2, 0.4, 0.8} 
\definecolor{mygreen}{rgb}{0.2, 0.6, 0.2} 
\definecolor{myteal}{RGB}{0,128,128}
\definecolor{mylavender}{HTML}{E6E6FA}

\newcommand{\redcell}{\cellcolor{red!30}}

\newcommand{\highlightred}[1]{\colorbox{red!30}{\strut\textbf{\textcolor{black}{#1}}}}
\newcommand{\highlightblue}[1]{\colorbox{blue!30}{\strut\textbf{\textcolor{black}{#1}}}}
\usepackage{amsmath,amssymb,amsfonts}

\usepackage[T1]{fontenc}

\usepackage[utf8]{inputenc}

\usepackage{microtype}

\usepackage{inconsolata}

\usepackage{graphicx}

\title{Marco-Bench-MIF: On Multilingual Instruction-Following Capability of Large Language Models}

\author{
\textbf{Bo Zeng\textsuperscript{1}},
\textbf{Chenyang Lyu\textsuperscript{1}}\thanks{Correspondence to lyuchenyang.lcy@alibaba-inc.com and wanglongyue.wly@alibaba-inc.com},
\textbf{Sinuo Liu\textsuperscript{1}},
\textbf{Mingyan Zeng},
\textbf{Minghao Wu\textsuperscript{1}},
\textbf{Xuanfan Ni\textsuperscript{1}}, \\
\textbf{Tianqi Shi\textsuperscript{1}},
\textbf{Yu Zhao\textsuperscript{1}}, 
\textbf{Yefeng Liu\textsuperscript{1}},
\textbf{Chenyu Zhu\textsuperscript{1}},
\textbf{Ruizhe Li\textsuperscript{2}},
\textbf{Jiahui Geng\textsuperscript{3}}, \\
\textbf{Qing Li\textsuperscript{3}}, 
\textbf{Yu Tong},
\textbf{Longyue Wang\textsuperscript{1}},
\textbf{Weihua Luo\textsuperscript{1}}, 
\textbf{Kaifu Zhang\textsuperscript{1}}
\\ \\
\textsuperscript{1}Alibaba International Digital Commerce,
\textsuperscript{2}University of Aberdeen,
\textsuperscript{3}MBZUAI \\
\\
}

\begin{document}
\maketitle
\begin{abstract}

Instruction-following capability has become a major ability to be evaluated for Large Language Models~(LLMs)~\cite{brown2020language_gpt3,openai2023gpt4,bai2023qwen}. However, existing datasets, such as IFEval~\cite{zhou2023instruction_ifeval,zeng2024evaluating_ifeval}, are either predominantly monolingual and centered on English or simply machine translated to other languages, limiting their applicability in multilingual contexts. In this paper, we present an carefully-curated extension of IFEval to a localized multilingual version named \texttt{Marco-Bench-MIF}, covering 30 languages with varying levels of localization. Our benchmark addresses linguistic constraints (e.g., modifying capitalization requirements for Chinese) and cultural references (e.g., substituting region-specific company names in prompts) via a hybrid pipeline combining translation with verification. Through comprehensive evaluation of 20+ LLMs on our \texttt{Marco-Bench-MIF}, we found that: (1) 25-35\% accuracy gap between high/low-resource languages, (2) model scales largely impact performance by 45-60\% yet persists script-specific challenges, and (3) machine-translated data underestimates accuracy by 7-22\% versus localized data. Our analysis identifies challenges in multilingual instruction following, including keyword consistency preservation and compositional constraint adherence across languages. Our \texttt{Marco-Bench-MIF} is available at \url{https://github.com/AIDC-AI/Marco-Bench-MIF}.
\end{abstract}

\section{Introduction}

Instruction-following datasets play a critical role in evaluating and fine-tuning Large Language Models~(LLMs)~\cite{geminiteam2023gemini,hurst2024gpt4o}, enabling them to better align with user intents. Among these, IFEval~\cite{zeng2024evaluating_ifeval} has emerged as a widely used benchmark for assessing instruction-following capabilities. However, its monolingual nature restricts its utility in multilingual and cross-lingual applications, where language-specific nuances and cultural contexts must be considered. {Existing datasets, including IFEval and Multi-IF~\cite{zhou2023instruction_ifeval,zeng2024evaluating_ifeval}, are either predominantly monolingual and centered on English or simply machine-translated to other languages, limiting their applicability in multilingual contexts. Machine-Translated (MT) data often fails to capture the linguistic and cultural 
subtle differences required for accurate evaluation, leading to underestimation of model performance in multilingual settings.} 

The construction of a multilingual instruction-following dataset should focus on adaptation of instructions, prompts, and constraints to diverse linguistic and cultural contexts. For example, tasks involving capitalization, such as \textit{``change all letters to uppercase''} should be modified for non-Latin scripts like Chinese and Greek, where such constraints are not applicable. Also tasks involving specific grammatical structures, such as \textit{``use the passive voice in your response''}, should be carefully adapted for languages like Japanese and Turkish, where passive constructions differ significantly from English. Similarly, prompts containing culturally specific references should be localized to align better with the target language and culture. For instance, the English prompt \textit{``Write a short story about a child celebrating Thanksgiving with their family''} can be adapted for the Indian context as \textit{``Write a short story about a child celebrating Diwali with their family, including the rituals of lighting diyas and sharing sweets.''} This adaptation not only preserves the instruction's structure but also ensures cultural relevance and linguistic appropriateness.

To address this gap, we extend IFEval to a multilingual version, encompassing 30 languages with varying levels of localization. {Our benchmark, named \texttt{Marco-Bench-MIF}, is carefully curated to address linguistic constraints (e.g., modifying capitalization requirements for Chinese) and cultural references (e.g., substituting region-specific company names in prompts). This is achieved through a hybrid pipeline that combines automated translation with two-round human verification to ensure quality and cultural relevance. Our localization process involved a combination of automated and manual methods. Initial translations and adaptations focus on content-specific localization, such as place names, events, and keywords. These outputs were then subjected to rigorous human evaluation and iterative refinement, including two rounds of manual review to ensure quality. Despite these efforts, challenges persisted, particularly in ensuring prompt adherence and accurately translating keywords. For example, in tasks requiring keyword frequency constraints, such as \textit{``ensure the word `sneaker' appears at least 10 times''}, translated keywords often required manual adjustment to maintain semantic consistency.

Our comprehensive evaluation of 20+ LLMs across 30 languages reveals critical insights into multilingual instruction-following capabilities. We observe that instruction-level accuracy consistently surpasses prompt-level metrics by 10-20\% across all models, highlighting persistent challenges in compositional instruction adherence. While model scale strongly correlates with performance (70B+ models achieve 45-60\% higher accuracy than 8B counterparts), even smaller models like Qwen2.5-7B attain 64.42\% strict instruction-level accuracy. Proprietary models (GPT-4o~\cite{hurst2024gpt4o}, Claude3.5-Sonnet~\cite{anthropic2024claude3}) significantly outperform open LLMs, with a 25-35\% accuracy gap in low-resource languages. The evaluation exposes substantial cross-lingual disparities: performance on high-resource languages (de, zh) is 75-85\% accuracy versus 50-60\% for low-resource languages (yo, ne), with script-specific challenges for some languages like Arabic (ar). Our analysis of localized versus machine-translated data demonstrates 7--22\% performance underestimation using translated data, showing the necessity for culturally-grounded evaluations. Our contributions in this paper are summarised as follows:

\begin{enumerate}
    \item  We extend IFEval to a multilingual benchmark, \texttt{Marco-Bench-MIF}, covering 30 languages with fine-grained localization to account for linguistic and cultural diversity.
    \item  We propose a systematic framework for adapting prompts, instructions, and constraints to diverse linguistic contexts, combining automated and manual localization methods, and analyze the challenges encountered during this process.
    \item  We conduct a comprehensive evaluation of 20+ LLMs on \texttt{Marco-Bench-MIF}, providing insights into multilingual instruction-following capabilities, including the impact of model scale, resource availability, and the limitations of machine-translated data.

\end{enumerate}

We believe this work represents a significant step forward in enabling robust and culturally-aware evaluation of LLMs instruction following ability across languages.

\section{Related Work}

Instruction-following datasets have become essential for evaluating LLMs. Early works such as GPT-3~\cite{brown2020language_gpt3} and GPT-4~\cite{openai2023gpt4} demonstrated the importance of this ability. However, existing benchmarks, such as IFEval~\cite{zeng2024evaluating_ifeval} and Multi-IF~\cite{he2024multiifbenchmarkingllmsmultiturn}, are either predominantly monolingual~(English) or just machine-translated data, limiting their applicability to multilingual and culturally diverse contexts.

Localization and cultural adaptation are key challenges as instructions must align with linguistic and cultural norms. For example, tasks involving capitalization are not relevant for non-Latin scripts like Chinese or Arabic. Cultural and linguistic appropriateness in LLM evaluation has also been explored, studies~\citep{liu2023cultural,romero2024cvqa, myung2024blend,chiu2024culturalbench} highlighted the importance of cultural alignment to ensure inclusivity and sensitivity in multilingual and multi-cultural settings. However, few works address the fine-grained challenges of adapting instructions for diverse languages and scripts. Manual intervention and human evaluation remain critical for ensuring quality and accuracy in multilingual datasets. Our work builds on these efforts by extending IFEval to a multilingual benchmark covering 30 languages. Unlike previous datasets, our extension emphasizes fine-grained localization to account for language-specific features and cultural differences. This includes adapting tasks that are not applicable to certain scripts and ensuring instructions are contextually relevant.

\section{Marco-Bench-MIF Dataset Construction}

We extend the IFEval benchmark into multilingual version through a systematic cross-lingual adaptation and a comprehensive pipeline. The framework consists of three main stages: preprocessing, translation localization, and post-processing. Our approach preserves instruction constraints while accommodating linguistic diversity across 30 languages from 6 major language families (Indo-European, Sino-Tibetan, Afro-Asiatic, etc.). Each language includes 541 instruction-response pairs, carefully constructed to ensure consistency and fidelity.

\subsection{Preprocessing}
The preprocessing stage involves categorizing and filtering the original IFEval dataset to prepare it for multilingual adaptation.

\subsubsection{Constraint Categorization}
We categorize the 541 English instruction-response pairs based on two major dimensions:
\begin{itemize}
    \item \textbf{Cardinality}: 1) \textit{Single-Constraint (SC)}: Instructions with a single requirement. 2) \textit{Multi-Constraint (MC)}: Instructions with two or more interdependent requirements.
    \item \textbf{Type}: 1) \textit{Expressive Constraints (EC)}: Dictate response structure or format. 2) \textit{Content Constraints (CC)}: Require specific information inclusion.
\end{itemize}

This categorization ensures a systematic adaptation process, starting with simpler constraints (SC+EC) before handling more complex ones (MC+CC). Table~\ref{tab:en_composition} shows the distribution of constraint types in the English benchmark. This graduated approach reduces error propagation risks, with simpler constraints (SC+EC) serving as anchors for more complex cases (MC+CC). 

\begin{table}[!htp]
\centering
\resizebox{\linewidth}{!}{
\begin{tabular}{llrrrr}
\toprule
\textbf{Cardinality} & \textbf{Const. Comp.} & \textbf{Count} & \textbf{Prop.} & \textbf{EC Freq.} & \textbf{CC Freq.} \\
\midrule
\multirow{2}{*}{SC} 
 & Pure EC & 109 & 20.1\% & 109 & 0 \\
 & Pure CC & 161 & 29.8\% & 0 & 161 \\
\cmidrule(lr){2-6}
 & SC Subtotal & 270 & 49.9\% & 109 & 161 \\
\midrule
\multirow{3}{*}{MC}
 & EC-only & 128 & 23.7\% & 128 & 0 \\
 & CC-only & 18 & 3.3\% & 0 & 18 \\
 & EC+CC Hybrid & 125 & 23.1\% & 125 & 125 \\
\cmidrule(lr){2-6}
 & MC Subtotal & 271 & 50.1\% & 253 & 143 \\
\midrule
\textbf{Total} & & 541 & 100.0\% & 362 & 304 \\
\bottomrule
\end{tabular}
}
\caption{Constraint Composition~(Const. Comp.) analysis of English benchmark (N=541)}
\label{tab:en_composition}
\end{table}

\begin{figure*}[!t]
    \centering
    \includegraphics[width=\linewidth]{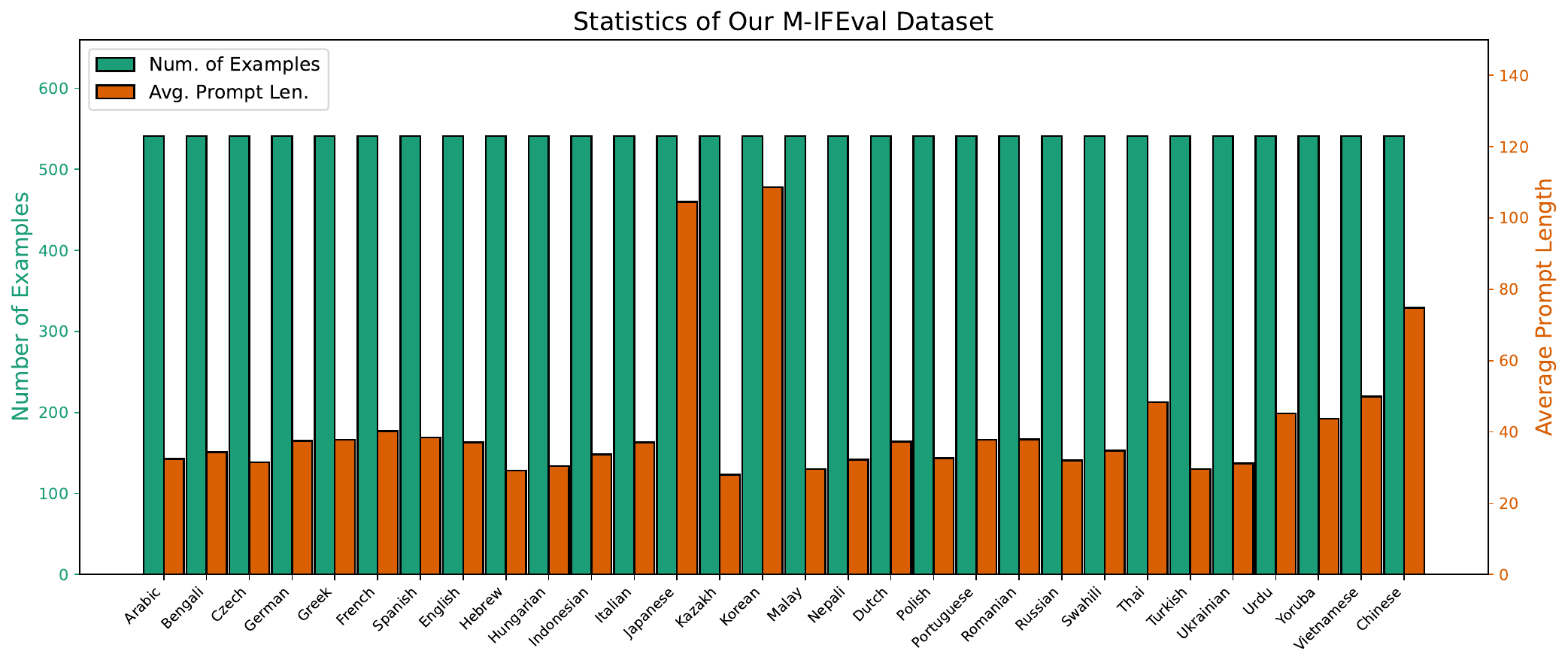}
    \caption{The number of examples and average prompt length for each language in our \texttt{Marco-Bench-MIF} dataset.}
    \label{fig:mifeval_stats}
\end{figure*}

\subsubsection{Data Filtering}
We filter the dataset to ensure high-quality instances for multilingual adaptation, the main principles are: 1). Removing ambiguous or overly complex instructions 2). Balancing the distribution of constraint types across languages 3). Ensuring consistency in instruction-response pairs.

\subsection{Translation and Localization}
After the pre-processing of the original instruction-following data, we move forward to expand it to multilingual via careful localization beyond simple translation. The translation localization stage adapts the dataset to 30 languages while preserving instruction constraints. Figure~\ref{fig:mifeval_stats} show the amount of examples and average length for the data in each language.

\subsubsection{Translation}
The translation process involves: 1) Initial translation using Google Translate. 2) Bilingual checks and manual corrections by professional translators. 3) Validation using LLMs for error correction. For five typologically diverse languages (Arabic, Spanish, Malay, Yoruba, Chinese), we create parallel corpus variants: machine-translated (MT) baselines and culturally localized versions. For the remaining 24 languages, we apply full cultural localization to SC+EC instances and strategic sampling for SC+CC and MC+EC cases.

\subsubsection{Localization}
Regarding further transforming the translated data to culturally localized version, we employ a three-step localization method:
\begin{itemize}
    \item \textbf{Lexical Substitution}: Replace culture-specific terms (e.g., names, locations) while maintaining constraint positions.
    \item \textbf{Topical Transposition}: Adapt scenario contexts to culturally familiar domains without altering constraint structures.
    \item \textbf{Pragmatic Restructuring}: Reformulate instructions using target-language rhetorical conventions under fixed constraints.
\end{itemize}

Cultural localization is achieved through a process guided by ten main sociolinguistic dimensions: historical contexts, social customs, lifestyle patterns, regional characteristics, geographical references, natural landscapes, traditional crafts, culinary traditions, entertainment forms, and national identity markers. We employ advanced LLMs perform constrained content regeneration across randomly selected thematic categories, preserving original task requirements while adapting cultural references. Then we use consensus for three LLMs to evaluate the localized outputs, any content receiving majority disapproval from three LLMs would receive mandatory human re-evaluation.

\subsection{Post-processing}

The dataset refinement process employs a multi-layered approach to ensure cross-lingual reliability through three interlocking mechanisms. For error reduction, we combine automated pattern detection with human check. The system specifically targets six prevalent translation failure points: keywords, exit statement, echo content, postscript consistency maintenance, case sensitivity adherence, and control of Latin character frequency in non-Latin script languages. We employ an LLM that generates initial outputs, which then undergo evaluation, with another LLM subsequently analyzing failure cases to distinguish between model capability limitations, instruction set defects, and evaluation logic flaws. Human reviewers then prioritize model-flagged cases for in-depth inspection of instruction adherence patterns and assessment code validity.

The evaluation framework undergoes systematic localization across 30 languages through four adaptation types: punctuation convention alignment, response language appropriateness verification, multi-section coherence validation, and constrained output requirement checking. Language selection balances regional prevalence, resource availability, and orthographic diversity, ensuring comprehensive coverage of global languages. This refinement process establishes Marco-Bench-MIF as a culturally-grounded benchmark for multilingual isntruction following.

\section{Evaluation on Marco-Bench-MIF}
\subsection{Evaluation Setup}
Our evaluation follows IFEval's  protocols~\cite{zhou2023instruction_ifeval} with additional adjustments. Basically, we prompt the LLMs with specific instructions $i$~(can be multiple instructions) to get a response $r$, we then analyze the response $r$ to verify whether the LLMs adhere the instruction $i$ we gave to it. Specifically, we employ rule-based verification and implement an additional one with text normalization. The \textbf{strict metric} is defined as following:
\[
E_{\text{strict}}(r, i) = \begin{cases} 
\text{1} & \text{if }r \text{ satisfies } i,\\[1mm]
\text{0} & \text{otherwise}
\end{cases}
\]

and \textbf{loose metric} with text normalization is:
\[
E_{\text{loose}}(r, i) = \bigvee_{t \in \mathcal{T}} \text{E}(\tau_t(r), i)
\]
where $\mathcal{T}$ indicating transformation operations (e.g., markdown removal, boundary adjustment) to reduce measurement variance while preserving instruction fidelity. If there is one text normalization can let $\tau_t(r)$ meet the instruction $i$ then we think is response is correct.

Moreover, within our evaluation methodology we have both the prompt and instruction level accuracy, which means:

\textbf{Prompt-level Accuracy}: This metric evaluates the model's ability to completely adhere to all verifiable instructions within a single prompt. For a response to be considered correct, the model must follow every instruction in its entirety, which is a metric that considers the models's ability for following multiple instructions together.

\textbf{Instruction-level Accuracy}: This measure assesses the model's compliance with each individual instruction within a prompt. A response is considered accurate on this level for the specific instruction if the model successfully follows the corresponding instructions provided, which is a more fine-grained metric for instruction-level performance.

\subsection{Evaluated LLMs}
We evaluate a diverse set of LLMs, encompassing both proprietary and open-source systems. The proprietary models include state-of-the-art systems including GPT-4o, Gemini, etc., while the open models include advanced LLMs with varying scales and architectures:

\paragraph{Proprietary Models.} The proprietary models evaluated in this study include \textbf{GPT} (GPT-4o, GPT-4o-Mini)~\cite{hurst2024gpt4o}, \textbf{Claude} (Claude3.5-Sonnet)~\cite{anthropic2024claude3}, \textbf{Gemini} (Gemini1.5-Flash)~\cite{geminiteam2023gemini}, and \textbf{Qwen-Proprietary} (Qwen-Plus, Qwen-Max)~\cite{bai2023qwen}. 

\paragraph{Open Models.} The open-source models include \textbf{Aya} (Aya-23: 8B, 35B; Aya-Expanse: 8B, 32B)~\cite{Ahmet2024ayamodelinstructionfinetuned}, \textbf{Gemma} (Gemma2: 9B, 27B)~\cite{gemmateam2024gemma2improvingopen}, \textbf{LLaMA} (LLaMA3.1-8B, LLaMA3.3-70B)~\cite{dubey2024llama3herdmodels}, \textbf{Qwen} (7B, 14B, 72B)~\cite{qwen2025qwen25technicalreport}, and \textbf{Mistral} (Mistral-Large, Ministral-8B)~\cite{jiang2023mistral7b}. Additionally, we include \textbf{EuroLLM-9B}~\footnote{\url{https://huggingface.co/utter-project/EuroLLM-9B-Instruct}}.

\subsection{Experimental Results}

\subsubsection{Overall Results}

\begin{table}[!t]
\centering
\resizebox{\linewidth}{!}{
\begin{tabular}{lccccc}
\toprule
Model & Prompt~(S) & Prompt~(L) & Inst.~(S) & \multicolumn{1}{c}{Inst.~(L)} & Avg. \\
\midrule
Ministral-8B & 21.74 & 24.49 & 46.45 & 49.72 & 35.60 \\
Aya23-8B & 23.46 & 26.03 & 46.89 & 49.23 & 36.40 \\
LLaMA3.1-8B & 25.48 & 28.36 & 48.98 & 51.66 & 38.62 \\
EuroLLM-9B & 32.02 & 35.33 & 56.42 & 59.32 & 45.77 \\
Aya23-35B & 34.95 & 38.79 & 57.24 & 60.43 & 47.85 \\
Aya-expanse-8B & 35.44 & 40.49 & 57.82 & 61.63 & 48.84 \\
Qwen2.5-7B & 42.99 & 47.43 & 64.42 & 68.02 & 55.72 \\
Aya-expanse-32B & 49.73 & 55.56 & 71.33 & 75.20 & 62.96 \\
Gemma2-9B & 51.97 & 54.53 & 72.16 & 73.89 & 63.14 \\
Qwen2.5-14B & 53.86 & 58.41 & 71.20 & 74.39 & 64.46 \\
Gemma2-27B & 58.86 & 61.35 & 77.21 & 78.78 & 69.05 \\
Mistral-Large & 62.66 & 67.16 & 79.02 & 81.96 & 72.70 \\
Qwen2.5-72B & 64.84 & 69.80 & \cellcolor{blue!30}80.80 & \cellcolor{blue!30}83.96 & 74.85 \\
LLaMA3.3-70B & \cellcolor{blue!30}67.42 & \cellcolor{blue!30}70.32 & 80.43 & 82.25 & \cellcolor{blue!30}75.11 \\
Qwen-Plus & 68.98 & 73.02 & 82.03 & 84.64 & 77.17 \\
Qwen-Max & 69.29 & 72.67 & 82.31 & 84.53 & 77.20 \\
GPT-4o-mini & 68.79 & 73.02 & 82.42 & 85.07 & 77.33 \\
Gemini1.5-flash & 71.40 & 75.88 & 84.17 & 86.95 & 79.60 \\
GPT-4o & 71.43 & 75.89 & 84.49 & 87.13 & 79.73 \\
Claude3.5-sonnet & \redcell 73.61 & \redcell 76.77 & \redcell 85.62 & \redcell 
87.71 & \redcell 80.93 \\
\bottomrule
\end{tabular}
}
\caption{Overall results of various LLMs on \texttt{Marco-Bench-MIF} with average accuracy, where the best performane of open models and proprietary model are marked in \highlightred{Red} and \highlightblue{Blue} respectively.}
\label{tab:overall_results}
\end{table}

The overall results of various LLMs on our \texttt{Marco-Bench-MIF} dataset are shown in Table~\ref{tab:overall_results}. The evaluation results across 30 languages show several interesting insights into the multilingual instruction-following capabilities of current LLMs.  First, a consistent trend across all models is that the instruction-level accuracy is substantially higher than the prompt-level accuracy (10-20\%). This indicates that although models may struggle to satisfy all instructions simultaneously when evaluated as a whole, they are often capable of executing individual instructions correctly. This effect is more obvious in smaller models like Ministral-8B (21.74 vs. 46.45 points difference), suggesting that compositional reasoning remains a key differentiator between model scales. Furthermore, the difference between strict and loose metrics underscores the influence of text normalization. The loose metric, which mitigates minor formatting or boundary discrepancies, typically produces higher accuracy scores than the strict metric (+3-5\%). This suggests that many models are close to correctly following instructions but occasionally falter due to superficial deviations in formatting, which do not fundamentally undermine the semantic intent of their responses.

\begin{table*}[!t]
\centering
\resizebox{\textwidth}{!}{%
\begin{tabular}{lccccccccccccccccccccccccccccccc}
\toprule
Model & ar\ & bn & cs & de & el & en & es\ & fr & he & hu & id & it & ja & kk & ko & ms\ & ne & nl & pl & pt & ro & ru & sw & th & tr & uk & ur & vi & yo\ & zh\ & Avg \\ 
\midrule
Ministral-8B & 21.8 & 23.1 & 33.8 & 54.6 & 32.4 & 53.0 & 53.4 & 52.2 & 30.3 & 33.9 & 36.8 & 50.0 & 43.7 & 23.8 & 33.8 & 33.2 & 15.7 & 41.9 & 34.5 & 51.0 & 32.6 & 49.0 & 24.1 & 23.5 & 32.0 & 33.5 & 17.3 & 33.5 & 20.6 & 53.0 & 35.7 \\
Aya23-8B & 50.3 & 26.2 & 37.5 & 45.7 & 33.9 & 44.3 & 43.7 & 45.1 & 43.7 & 23.7 & 37.6 & 46.2 & 45.3 & 13.7 & 41.9 & 32.5 & 22.5 & 38.1 & 42.0 & 45.9 & 34.7 & 40.4 & 21.9 & 36.2 & 30.8 & 40.7 & 27.4 & 44.3 & 15.9 & 39.6 & 36.4 \\
LLaMA3.1-8B & 24.2 & 39.7 & 40.6 & 51.0 & 32.1 & 71.4 & 55.4 & 53.9 & 31.0 & 36.2 & 44.7 & 49.7 & 29.8 & 26.1 & 23.8 & 41.6 & 22.1 & 45.8 & 37.9 & 53.3 & 44.4 & 33.1 & 32.0 & 35.9 & 36.0 & 27.0 & 22.1 & 46.0 & 16.3 & 33.5 & 37.9 \\
LLaMA3-8B & 28.7 & 23.1 & 51.8 & 50.5 & 46.2 & 70.8 & 63.1 & 58.8 & 31.8 & 48.4 & 54.9 & 55.2 & 25.2 & 29.2 & 24.3 & 41.7 & 17.4 & 59.6 & 35.7 & 55.5 & 54.1 & 37.1 & 30.2 & 24.8 & 47.1 & 30.7 & 19.0 & 30.4 & 16.4 & 29.8 & 39.7 \\
Qwen2-7B & 50.6 & 43.6 & 44.1 & 54.3 & 37.2 & 57.9 & 55.8 & 56.9 & 29.8 & 34.9 & 49.5 & 55.5 & 53.5 & 29.0 & 49.3 & 40.7 & 34.2 & 51.4 & 46.1 & 55.7 & 42.1 & 51.4 & 30.5 & 51.3 & 41.6 & 44.3 & 42.0 & 49.8 & 19.0 & 59.6 & 45.4 \\
EuroLLM-9B & 59.2 & 31.4 & 48.7 & 63.7 & 51.0 & 63.5 & 63.1 & 62.2 & 34.9 & 45.6 & 40.2 & 60.7 & 51.7 & 24.0 & 50.8 & 37.3 & 25.6 & 53.1 & 53.2 & 60.2 & 47.2 & 57.1 & 24.6 & 34.4 & 43.4 & 49.5 & 29.0 & 38.7 & 21.7 & 59.9 & 46.2 \\
Aya23-35B & 59.0 & 33.1 & 50.4 & 60.6 & 49.4 & 60.0 & 57.0 & 59.3 & 57.5 & 33.8 & 51.7 & 59.2 & 58.7 & 18.9 & 54.0 & 44.9 & 31.6 & 51.4 & 51.5 & 58.3 & 46.4 & 51.6 & 27.6 & 39.9 & 47.9 & 49.9 & 36.9 & 56.8 & 16.1 & 52.3 & 47.5 \\
Aya-expanse-8B & 61.7 & 34.4 & 54.5 & 62.1 & 51.8 & 65.1 & 64.7 & 63.3 & 58.7 & 31.5 & 51.8 & 62.5 & 56.3 & 18.2 & 54.6 & 43.5 & 11.8 & 55.7 & 54.2 & 60.2 & 49.4 & 56.0 & 31.9 & 40.7 & 47.3 & 53.5 & 39.9 & 55.5 & 21.0 & 63.5 & 49.2 \\
LLaMA3-70B & 41.6 & 33.3 & 70.0 & 57.8 & 67.1 & 82.0 & 76.5 & 70.5 & 39.0 & 68.2 & 72.1 & 63.9 & 35.4 & 46.1 & 31.7 & 58.1 & 28.3 & 74.6 & 44.3 & 69.3 & 69.4 & 50.2 & 41.1 & 32.3 & 64.8 & 43.4 & 33.5 & 38.2 & 28.9 & 40.5 & 52.4 \\
LLaMA3.1-70B & 32.4 & 61.9 & 59.2 & 68.1 & 49.1 & 81.6 & 69.0 & 64.4 & 57.9 & 59.2 & 58.7 & 64.8 & 44.6 & 45.9 & 31.2 & 58.5 & 49.7 & 61.1 & 52.9 & 64.6 & 61.3 & 45.4 & 52.8 & 50.6 & 50.8 & 38.6 & 38.6 & 61.8 & 28.9 & 44.8 & 53.6 \\
Qwen2.5-7B & 55.4 & 56.9 & 58.9 & 63.6 & 49.4 & 73.7 & 67.8 & 68.8 & 35.0 & 53.5 & 66.6 & 68.0 & 60.2 & 35.7 & 51.2 & 59.5 & 43.5 & 62.7 & 52.5 & 69.0 & 61.9 & 60.6 & 32.9 & 57.3 & 50.9 & 56.2 & 43.4 & 60.5 & 19.8 & 70.5 & 55.5 \\
Aya-expanse-32B & 74.6 & 49.9 & 66.5 & 75.3 & 65.8 & 74.3 & 73.9 & 72.9 & 70.8 & 53.8 & 67.5 & 75.1 & 72.0 & 32.4 & 69.8 & 64.8 & 41.5 & 69.1 & 66.5 & 73.9 & 62.2 & 69.3 & 39.8 & 52.0 & 64.5 & 66.1 & 56.2 & 66.6 & 19.1 & 75.4 & 62.7 \\
Gemma2-9B & 67.6 & 59.0 & 63.7 & 73.4 & 61.1 & 74.4 & 72.6 & 72.5 & 65.3 & 60.9 & 66.2 & 72.0 & 62.9 & 42.5 & 42.5 & 66.0 & 57.5 & 68.7 & 61.4 & 72.4 & 62.5 & 67.2 & 62.5 & 62.4 & 59.8 & 63.6 & 60.2 & 64.0 & 30.1 & 68.8 & 62.8 \\
Qwen2.5-14B & 70.2 & 67.3 & 62.5 & 72.4 & 61.0 & 83.4 & 74.7 & 75.3 & 39.4 & 59.7 & 68.5 & 74.1 & 70.9 & 43.3 & 66.7 & 64.7 & 59.7 & 68.4 & 58.1 & 75.9 & 66.0 & 69.5 & 45.1 & 70.2 & 57.4 & 65.0 & 63.4 & 68.5 & 21.3 & 74.8 & 63.9 \\
Qwen2-72B & 74.0 & 68.9 & 72.0 & 79.6 & 65.7 & 80.4 & 79.4 & 80.5 & 39.0 & 65.2 & 74.6 & 79.9 & 74.2 & 44.2 & 69.2 & 71.6 & 56.4 & 75.0 & 65.8 & 79.2 & 72.8 & 74.8 & 48.4 & 70.7 & 60.6 & 66.7 & 67.6 & 72.5 & 19.5 & 78.0 & 67.5 \\
Gemma2-27B & 70.9 & 64.7 & 68.2 & 74.5 & 66.8 & 81.0 & 76.8 & 76.0 & 71.8 & 69.1 & 70.7 & 74.6 & 66.3 & 57.3 & 69.2 & 67.9 & 66.1 & 73.5 & 64.8 & 73.2 & 66.7 & 71.2 & 68.7 & 64.9 & 68.8 & 66.8 & 64.1 & 67.3 & 34.3 & 73.2 & 68.3 \\
Mistral-Large & 75.2 & 65.2 & 77.3 & 83.4 & 71.2 & 81.9 & 82.0 & 82.3 & 71.3 & 73.6 & 77.2 & 82.5 & 73.6 & 49.2 & 76.5 & 73.0 & 63.6 & 79.8 & 68.7 & 79.8 & 73.4 & 75.7 & 54.6 & 69.3 & 71.7 & 71.8 & 68.7 & 75.4 & 17.8 & 78.3 & 71.5 \\
LLaMA3.3-70B & 61.9 & \cellcolor{blue!30}79.0 & 75.7 & 82.3 & 70.2 & \cellcolor{blue!30}90.5 & 84.6 & 83.0 & \cellcolor{blue!30}78.5 & \cellcolor{blue!30}75.4 & 77.8 & \cellcolor{blue!30}83.6 & \cellcolor{blue!30}76.8 & \cellcolor{blue!30}64.2 & 62.5 & \cellcolor{blue!30}78.8 & \cellcolor{blue!30}73.8 & \cellcolor{blue!30}80.8 & \cellcolor{blue!30}68.7 & 81.0 & 76.8 & 72.2 & \cellcolor{blue!30}73.3 & \cellcolor{blue!30}75.0 & 70.2 & 54.7 & \cellcolor{blue!30}73.7 & 77.8 & \cellcolor{blue!30}37.5 & 74.2 & 73.8 \\
Qwen2.5-72B & \cellcolor{blue!30}77.8 & 76.2 & \cellcolor{blue!30}79.9 & \cellcolor{blue!30}82.5 & \cellcolor{blue!30}75.4 & 85.7 & \cellcolor{blue!30}84.7 & \cellcolor{blue!30}83.5 & 51.4 & 73.8 & \cellcolor{blue!30}82.5 & 81.8 & 76.1 & 59.0 & \cellcolor{blue!30}74.2 & 78.3 & 72.2 & 80.7 & 67.5 & \cellcolor{blue!30}85.0 & \cellcolor{blue!30}77.0 & \cellcolor{blue!30}78.2 & 55.9 & 74.7 & \cellcolor{blue!30}74.9 & \cellcolor{blue!30}75.6 & 68.8 & \cellcolor{blue!30}78.9 & 27.6 & \cellcolor{blue!30}81.8 & \cellcolor{blue!30}74.0 \\
GPT-4o-mini & 79.7 & 78.1 & 77.0 & 84.1 & 75.8 & 84.9 & 83.2 & 81.7 & 80.3 & 77.6 & 76.5 & 81.7 & 80.0 & 70.0 & 79.2 & 77.3 & 76.3 & 81.5 & 69.5 & 83.3 & 76.2 & 77.5 & 76.6 & 80.2 & 75.2 & 75.3 & 77.6 & 81.0 & 51.4 & 81.5 & 77.7 \\
Qwen-Max & 81.8 & 75.7 & 81.0 & 85.8 & 73.5 & 87.4 & 84.1 & 85.0 & 77.4 & 78.2 & 83.9 & 85.6 & 81.0 & 66.7 & 80.9 & 80.9 & 69.8 & 81.9 & \redcell 74.9 & 84.1 & 80.2 & \redcell 81.5 & 66.8 & \redcell 81.2 & 80.3 & 77.7 & 78.3 & 81.2 & 39.7 & 80.6 & 78.2 \\
Qwen-Plus & 84.2 & 77.6 & \redcell 84.2 & \redcell 87.3 & 75.6 & 89.5 & \redcell 87.7 & \redcell 87.6 & 72.6 & 78.6 & 84.8 & \redcell 85.9 & \redcell 81.3 & 63.9 & 81.2 & 82.2 & 73.2 & 84.0 & 73.5 & \redcell 86.0 & \redcell 82.3 & 79.9 & 60.1 & 80.3 & 79.5 & 78.2 & 75.9 & \redcell 83.8 & 24.8 & \redcell 83.7 & 78.3 \\
GPT-4o & 81.2 & 79.7 & 79.7 & 84.2 & 79.0 & 87.8 & 84.2 & 85.5 & 80.9 & 78.6 & 82.9 & 84.3 & 79.4 & 74.8 & 79.3 & 79.4 & 78.8 & 84.8 & 73.0 & 85.0 & 79.6 & 78.6 & 79.3 & 78.9 & 79.2 & 78.1 & 79.8 & 81.1 & 59.1 & 83.6 & 80.0 \\
Gemini1.5-flash & \redcell 84.9 & 78.2 & 80.7 & 85.6 & 77.8 & 89.5 & 84.2 & 86.0 & 81.3 & 78.6 & 83.4 & 85.1 & 77.8 & 66.4 & 79.3 & \redcell 83.8 & 78.7 & 83.1 & 72.6 & 84.2 & 78.0 & 80.4 & 80.0 & 79.6 & 78.8 & \redcell 78.8 & 78.6 & 83.5 & 60.5 & 81.6 & 80.0 \\
Claude3.5-sonnet & 82.4 & \redcell 81.5 & 80.9 & 84.4 & \redcell 79.3 & \redcell 90.3 & 87.0 & 87.3 & \redcell 82.3 & \redcell 82.7 & \redcell 85.2 & 84.9 & 80.2 & \redcell 75.6 & \redcell 82.2 & 82.4 & \redcell 80.2 & \redcell 84.9 & 74.5 & 85.0 & 81.0 & 80.9 & \redcell 81.3 & 81.0 & \redcell 83.3 & 78.0 & \redcell 80.6 & 82.2 & \redcell 62.3 & 82.4 & \redcell 81.5 \\
Avg & 62.1 & 56.3 & 64.0 & 70.7 & 59.9 & 76.2 & 72.3 & 71.8 & 56.5 & 59.0 & 65.9 & 70.7 & 62.3 & 44.8 & 58.4 & 61.7 & 50.0 & 67.7 & 58.6 & 70.8 & 63.1 & 63.6 & 49.7 & 57.9 & 59.9 & 58.5 & 53.7 & 63.2 & 29.2 & 65.8 & 60.8 \\

\bottomrule
\end{tabular}%
}
\caption{Overall results for our evaluated LLMs divided by language, where the best performane of open models and proprietary model are marked in \highlightred{Red} and \highlightblue{Blue} respectively.}
\label{tab:results_per_language}
\end{table*}

Another observation is the trend in performance as model capacity increases, we observe a strong correlation between model scale and performance, with larger models (70B+ parameters) achieving 45-60\% higher absolute accuracy than 8B LLMs. However, this scaling effect is not as obvious as it is in prompt-level for instruction-level metrics, where even smaller models like Qwen2.5-7B achieve 64.42\% strict accuracy, suggesting that basic instruction understanding can be achieved at that scales. Moreover, proprietary LLMs all exhibit strong performance against open LLMs, which demonstrates the superiority of these models and that there is still large room for improvement for open models.

\subsubsection{Results per Language}

We split the evaluation results by the language in our \texttt{Marco-Bench-MIF} dataset, the results are shown in Table~\ref{tab:results_per_language}. The results show three critical patterns in multilingual instruction following capability of LLMs. First, the performance of LLMs is largely depended on the availability of the language resources: High-resource European (de, fr) and East Asian languages (zh, ja) achieve 75-85\% accuracy across top models, while low-resource languages (yo, ne, kk) performance is at 50-60\% even for Claude3.5-sonnet. This 25-35 point gap persists across difference LLM scales, suggesting that current state-of-the-art LLMs still struggle for low-resource languages at instruction-following.

\begin{figure}[!t]
    \centering
    \includegraphics[width=\linewidth]{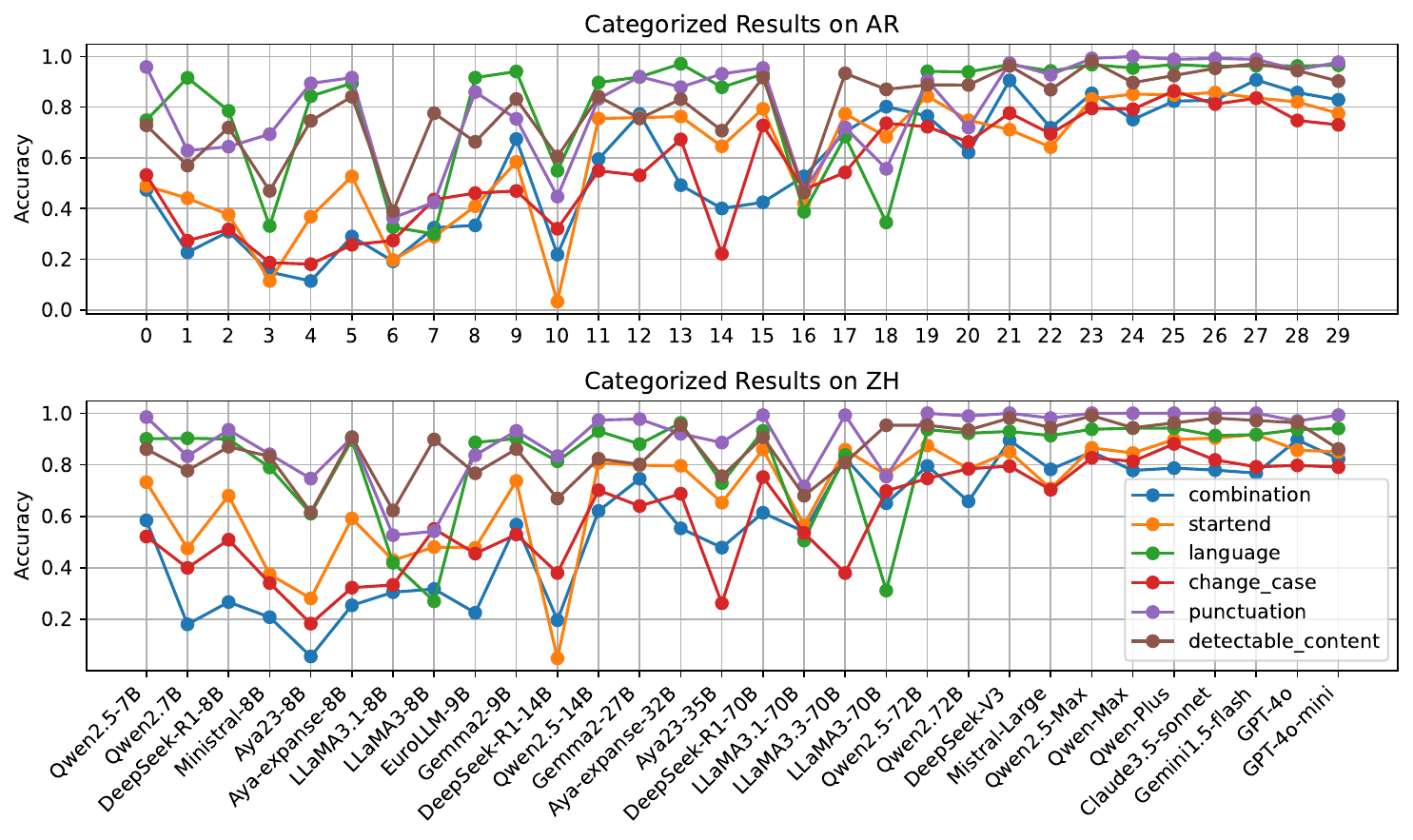}
    \caption{Results of LLMs on Arabic (Ar) and Chinese~(Zh) data divided by categories.}
    \label{fig:results_ar_zh}
\end{figure}

Second, \textbf{script-specific capabilities} significantly impact performance. Most LLMs exhibit 15-20 point disparities between languages using the same script family (e.g., Romance languages es/pt vs. Romanian ro), indicating character-level encoding challenges. Right-to-left scripts (ar, he) show particular sensitivity, with LLaMA3.3-70B scoring 78.5 in Hebrew versus 54.7 in Urdu (ur) despite similar pretraining data sizes. We see an interesting point from the results that while multilingual-specialized models like Aya-expanse-32B achieve best-in-class performance for mid-resource languages (hi:56.2, sw:39.8), general-purpose models like Qwen2.5-72B outperform them in 22/30 languages through scale advantages.  We identify three \textbf{language response types}: 1). \textit{Scale-sensitive} languages (ru, tr) where 70B models achieve almost 20 point gains over 8B LLMs 2). \textit{Model-dependent} languages (he, ja) showing 15-30 point variations between equally-sized models 3). \textit{Extremely-hard} languages (yo, kk) where even GPT-4o scores less than 60\% accuracy Moreover, \textbf{cross-lingual interference} patterns appear in related language pairs. For Turkic languages, Turkish (tr) performance strongly predicts Kazakh (kk) accuracy (r=0.89), suggesting models transfer capabilities within language families. However, this correlation does not hold for Slavic languages, where Ukrainian (uk) shows unexpectedly low scores relative to Russian (ru) across all models.

We further plot the categorized results on two languages with different scripts (Latin vs non-Latin) of our \texttt{Marco-Bench-MIF}, results shown in Figure~\ref{fig:results_ar_zh}. There are some interesting findings and patterns. Closed-source models like Qwen-Max and GPT-4o maintain robust performance across both languages (less than 5\% relative difference in \textit{detectable\_content}), we observe significant language-specific degradation in smaller open-source models - particularly for structural constraints like \textit{startend} ($\Delta$=28\% for Qwen2-7B between zh/ar). Chinese processing shows stronger format preservation (avg. 89\% vs. 82\% in \textit{detectable\_format}) while Arabic models better handle \textit{language} identification (94\% vs. 91\% avg.), suggesting script-specific pattern learning biases. The non-Latin script challenge emerges clearly in \textit{change\_case} tasks, where all models show 12-15\% performance drops for Arabic compared to Chinese, revealing limitations in script-invariant processing. We identify a critical cross-lingual transfer gap: Mistral-Large achieves 87\%/94\% \textit{detectable\_content} for ar/zh respectively, while LLaMA3-70B shows dramatic variance (35\% vs. 95\%), indicating inconsistent multilingual generalization. Punctuation handling proves most language-agnostic ($\Delta$<3\% for top models), suggesting structural linguistic features rather than script properties dominate this capability. Our findings emphasize that effective multilingual instruction following requires separate optimization beyond simple vocabulary extension, particularly for combinatorial constraints in right-to-left scripts ($\Delta$=19\% for \textit{combination} tasks ar vs. zh).

\begin{figure}[!t]
    \centering
    \includegraphics[width=\linewidth]{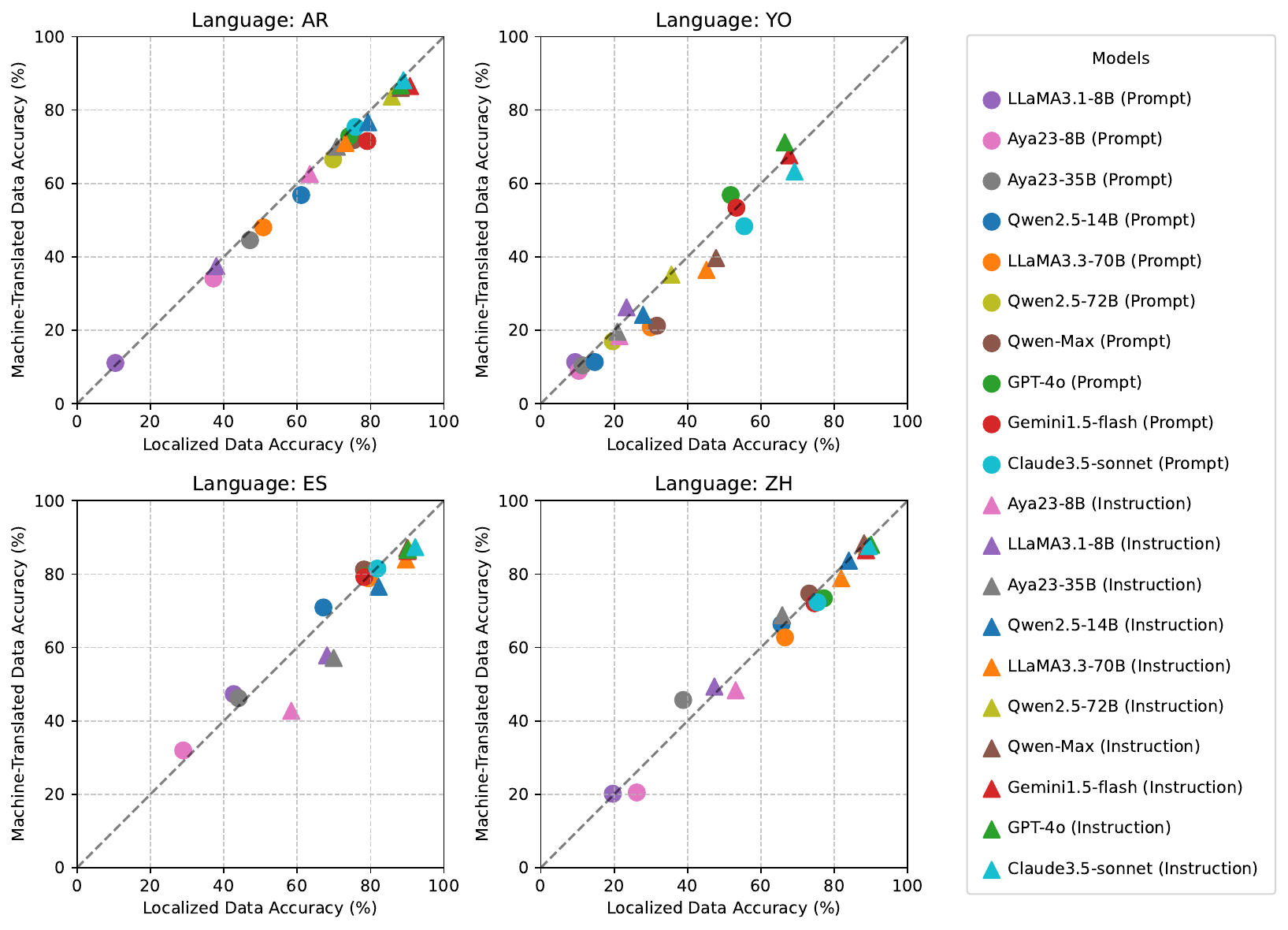}
    \caption{Comparison between the performance (average accuracy of \textit{Strict} and \textit{Loose} mode for both instruction and prompt-level) of various LLMs on the localized data vs machine-translated data.}
    \label{fig:results_mt_vs_local}
\end{figure}

\subsubsection{Responding in Specific Language}
We study an interesting aspect of multilingual instruction following capability - let LLMs respond in specified language. This sub-task directly reflects the text generation capability of LLMs in various language. The results are shown in Table~\ref{tab:response_lang}. Results show that multilingual-enhanced LLMs: Aya-expanse outperform proprietary models regardless of size, with the Aya-expanse-32B variant (95.69) surpassing most proprietary systems. This demonstrates the effectiveness of targeted multilingual enchancement. Gemma models (both 9B and 27B) also achieve comparable performance (91+) to top proprietary models. We can see that even top LLMs like GPT-4o and Claude still can not fully follow the instruction that respond in specified language (5\% wrong outputs).

\begin{table}[!t]
\centering
\resizebox{0.6\linewidth}{!}{
\begin{tabular}{lc}
\toprule
Model & Response Language \\
\midrule
\multicolumn{2}{l}{\textbf{Open Models(< 10B)}} \\
\quad LLaMA3.1-8B       & 58.34 \\
\quad Aya23-8B           & 65.57 \\
\quad Ministral-8B       & 66.03 \\
\quad Aya-expanse-8B     & 75.56 \\
\quad Qwen2.5-7B         & 78.95 \\
\quad EuroLLM-9B         & 79.19 \\
\quad Gemma2-9B          & 91.82 \\
\midrule
\multicolumn{2}{l}{\textbf{Open Models (> 10B)}} \\
\quad Aya23-35B          & 68.88 \\
\quad Qwen2.5-14B        & 79.69 \\
\quad Gemma2-27B         & 91.88 \\
\quad Aya-expanse-32B    & 95.69 \\
\quad LLaMA3.3-70B       & 82.38 \\
\quad Qwen2.5-72B        & 91.02 \\
\midrule
\multicolumn{2}{l}{\textbf{Proprietary Models}} \\
\quad Mistral-Large      & 90.67 \\
\quad Qwen-Plus          & 92.22 \\
\quad Qwen-Max           & 93.20 \\
\quad Claude3.5-sonnet   & 94.03 \\
\quad Gemini1.5-flash    & 94.08 \\
\quad GPT-4o-mini        & 95.09 \\
\quad GPT-4o             & 95.43 \\
\bottomrule
\end{tabular}
}
\caption{Results of the \textit{Response Language} instruction (average accuracy of \textit{Strict} and \textit{Loose} mode).}
\label{tab:response_lang}
\end{table}

\subsubsection{Effect of Translated and Localised Data}

\begin{table*}[!h]
\centering
\resizebox{0.9\textwidth}{!}{
\begin{tabular}{p{0.4\textwidth}p{0.35\textwidth}p{0.4\textwidth}}
\toprule
\textbf{Input Instruction} & \textbf{Open Models} & \textbf{Proprietary Models} \\
\midrule
\textbf{Chinese:}\newline
\begin{CJK*}{UTF8}{gbsn}
讲一个包含单词"星期四"和"合并"的笑话，但只能使用斯瓦希里语，不允许使用其他语言。
\end{CJK*}
\textit{\small (Tell a joke containing the words "Thursday" and "merge" in Swahili only, no other languages allowed.)} & 
\textbf{Aya-expanse-32B:}\newline
\begin{CJK*}{UTF8}{gbsn}
Sijambo, ni sehemu ambako tunapokaa tuck {\bfseries kulaungiana} mikosi ya juma la saba na {\bfseries juma la nne}...
\end{CJK*}
\textit{\small (Hello, it's a place where we stay to merge the misfortunes of Saturday and Thursday...)} \newline
\textcolor{red}{\textbf{\textit{correct}}}& 
\textbf{GPT-4o:}\newline
\begin{CJK*}{UTF8}{gbsn}
为什么星期四和星期五不能合并？因为星期四还在努力工作，而星期五已经开始庆祝周末了！
\end{CJK*}
\textit{\small (Why can't Thursday and Friday merge? Because on Thursday you're still working hard, but on Friday you've already started celebrating the weekend!)}\newline
\textcolor{red}{\textbf{\textit{wrong}}} \\
\hline

\textbf{Spanish:}\newline
Reescribe la frase: 'Lancé mi copla al viento de Levante'. Escribe toda tu respuesta entre comillas dobles.
\textit{\small (Rewrite the phrase: 'I threw my verse to the Levante wind'. Write your entire answer in double quotes.)} & 
\textbf{LLaMA3.3-70B:}\newline
"Dejo que mi canción se escuche llevada por el viento de Levante"
\textit{\small (I let my song be heard carried by the Levante wind)} \newline
\textcolor{red}{\textbf{\textit{wrong}}}& 
\textbf{Claude-3.5:}\newline
"Al viento de Levante arrojé mi copla" 
\textit{\small (To the Levante wind I threw my verse)}\newline
\textcolor{red}{\textbf{\textit{wrong}}} \\
\hline

\textbf{Malay:}\newline
Boleh buat jadual lawatan 5 hari ke Istana Maimoon yang mengandungi tepat 3 titik peluru dalam format senarai menurun dengan huruf kecil dan nota di akhir bermula dengan P.S.?
\textit{\small (Can you make a 5-day visit schedule to Maimoon Palace containing exactly 3 bullet points in a descending list format with lowercase letters and a note at the end starting with P.S.? Please answer in English.)} & 
\textbf{Qwen2.5-72B:}\newline
\textit{Pulau Riau emas terbit}\newline
\textit{Laut lebar burung terbang}\newline
\textit{Rindang kelapa berjajar}\newline
\textit{Harapan bersama di Nusantara}
\textit{\small (Riau Island golden sunrise / Wide sea birds flying / Palm trees in rows / Hope together in Nusantara)} \newline
\textcolor{red}{\textbf{\textit{wrong}}}& 
\textbf{Gemini1.5-flash:}\newline
\textit{Pulau Bintan indah sekali}\newline
\textit{Lautnya biru pasir putih}\newline
\textit{Harta budaya Melayu asli}\newline
\textit{Tersebar luas di Nusantara}
\textit{\small (Bintan Island is beautiful / Blue sea white sand / Original Malay cultural heritage / Spread widely in Nusantara)} \newline
\textcolor{red}{\textbf{\textit{wrong}}}\\
\bottomrule
\end{tabular}
}
\caption{Examples of multilingual instruction-following capabilities across different models. English translations are provided in italics below each non-English text.}
\label{tab:multilingual_examples}
\vspace{-1em}
\end{table*}

We also conduct experiments investigating the effects of localized data versus machine-translated data, the results are shown in Figure~\ref{fig:results_mt_vs_local}.  The results clearly demonstrate that evaluations based exclusively on MT data can be misleading. In high-resource languages such as Arabic and Chinese, the differences appear minimal, but for low-resource languages like Yoruba the gap is more larger (prompt-level: avg 22.13\% localized vs. 19.77\% MT). For example, the performance of Claude3.5-sonnet in Yoruba is largely underestimated by 7.1\% when evaluated using MT data.
These discrepancies indicate that MT-data can not precisely capture the full complexity of instruction following—especially when it comes to culturally and linguistically fine-grained inputs. As task complexity increases, the inadequacy of MT data becomes even more evident, suggesting that localized data is crucial for a precise evaluation of LLMs' instruction-following abilities.

\subsection{Case Study}

We demonstrate some test cases and model outputs from our \texttt{Marco-Bench-MIF} dataset in Table~\ref{tab:multilingual_examples}. The three representative cases provide some concrete observation of LLMs behavior when facing localized multilingual instructions. The Chinese-Swahili joke generation exposes fundamental language switching limitations, where GPT-4o defaults to Chinese despite explicit instructions, while Aya-expanse-32B successfully maintains Swahili output through specialized multilingual embeddings. Both LLaMA3.3-70B and Claude-3.5 fail the Spanish rewriting task's structural requirement (strict quote enclosure), demonstrating that formatting constraints remain under-addressed in current multilingual training paradigms. The Malay schedule generation reveals deeper instruction comprehension failures - proprietary models generate irrelevant poetry instead of bullet-point schedules, suggesting overfitting to frequent Malay literary patterns rather than practical task requirements. These failures persist across model scales and architectures, indicating systemic gaps in non-English instruction following that correlate with training data priorities rather than model capacity. The cases highlight three underdeveloped capabilities of LLMs: 1) cross-lingual constraint propagation, 2) non-English structural formatting, and 3) culture-specific task understanding.

\section{Conclusion}

Our work presents \texttt{Marco-Bench-MIF}, a comprehensive multilingual instruction-following benchmark covering 30 languages with localized adaptations. Through extensive evaluation, we identify key challenges in multilingual instruction following, including \textit{formatting fragility}, \textit{compositional reasoning limits}, and \textit{structural-task disparity}. Our results demonstrate that while large-scale models achieve strong performance in high-resource languages, significant gaps remain for low-resource languages and complex structural constraints. 
These findings highlight the need for more robust evaluation frameworks that account for linguistic diversity and task complexity.
Specialized models like Aya-expanse-32B in certain multilingual instructions suggests that targeted architectural improvements can complement scaling strategies. However, the persistent performance gaps in low-resource languages and fine-grained cultural-related tasks indicate that data quality and diversity remain critical bottlenecks. 
Future research will focus on expanding \texttt{Marco-Bench-MIF} to include more languages particularly those with unique typological features, will provide a more comprehensive understanding of multilingual instruction-following capabilities. We also plan to include more instruction types to better reflect real-world scenarios.

\section*{Limitations}

Our work has several limitations that highlight directions for improvement. First, while our \texttt{Marco-Bench-MIF} covers 30 languages, it underrepresents languages with non-Latin scripts (e.g., Ethiopic, Cherokee) and dialectal variations (e.g., Arabic dialects), which limits investigation into instruction with more diverse scripts. Second, our cultural localization focuses on surface-level adaptations (e.g., date formats) with a few fine-grained cultural pragmatics (e.g., politeness strategies in Japanese), potentially missing more  deeper instruction-following failures. Moreover, while we mitigated translation artifacts through human evaluation, residual biases from automated localization (e.g., GPT-4’s tendency to formal register) may still exist in certain languages.
\textbf{Finally}, our evaluation focuses on static prompts rather than interactive instruction refinement, limiting insights into models’ ability to recover from errors in dynamic conversations.

\section*{Ethics and Broader Impact}

In developing \texttt{Marco-Bench-MIF}, a multilingual instruction-following benchmark covering 30 languages, we prioritize ethical considerations in dataset creation, model evaluation, and broader societal impacts. Our localization process combines automated translation with human verification to ensure cultural sensitivity and linguistic appropriateness, avoiding stereotypes or misrepresentations. The dataset will be made publicly available with detailed documentation to ensure transparency and reproducibility, while clear usage guidelines will mitigate potential misuse, such as deploying LLMs in harmful applications. We also recognize the environmental impact of computational resources used in this work and advocate for energy-efficient methods in NLP research. We are committed to responsible AI development and have taken steps to ensure fairness, inclusivity, and respect for human contributors. Human annotators and translators were compensated fairly and provided informed consent, with their privacy and rights upheld throughout the project. 

\bibliography{custom}

\appendix

\section{Appendix}
\label{sec:appendix}

\subsection{Results per Category}
\label{appendix:subsec:results_per_category}

\begin{table}
\centering
\resizebox{\linewidth}{!}{
\begin{tabular}{cccc}
\toprule
Model & no\_comma & end\_checker & quotation \\ 
\midrule
Aya23-8B & 28.08 & 26.80 & 25.61 \\
Ministral-8B & 41.15 & 27.91 & 21.66 \\
LLaMA3.1-8B & 44.73 & 32.77 & 28.88 \\
EuroLLM-9B & 41.64 & 38.30 & 37.75 \\
Aya-expanse-8B & 24.57 & 44.06 & 54.68 \\
Aya23-35B & 46.21 & 40.92 & 56.38 \\
Qwen2.5-7B & 67.33 & 46.22 & 56.09 \\
Aya-expanse-32B & 46.75 & 52.54 & 74.00 \\
Gemma2-9B & 68.67 & 63.92 & 64.34 \\
Qwen2.5-14B & 67.03 & 69.16 & 68.05 \\
Mistral-Large & 86.48 & 70.62 & 64.68 \\
Gemma2-27B & 87.84 & 66.92 & 78.92 \\
GPT-4o-mini & 87.44 & 72.00 & 85.26 \\
Qwen2.5-72B & 87.80 & 73.39 & 83.86 \\
Qwen-Plus & 90.36 & 72.44 & 83.09 \\
Qwen-Max & 91.84 & 73.50 & 81.75 \\
GPT-4o & 86.50 & 75.23 & 89.42 \\
LLaMA3.3-70B & 92.07 & 82.16 & 83.27 \\
Gemini1.5-flash & 93.23 & 77.90 & 89.73 \\
Claude3.5-sonnet & 95.56 & 77.89 & 94.12 \\
\bottomrule
\end{tabular}
}
\caption{Format style Instruction Following Results}
\end{table}

\begin{table*}
\centering
\resizebox{\textwidth}{!}{
\begin{tabular}{ccccccccc}
\toprule
Model & repeat\_prompt & two\_responses & number\_placeholders & postscript & existence & forbidden\_words & frequency & letter\_frequency \\
\midrule
Aya23-8B & 6.86 & 10.02 & 44.66 & 61.18 & 44.86 & 68.70 & 41.00 & 47.19 \\
EuroLLM-9B & 7.86 & 32.25 & 41.33 & 75.39 & 54.95 & 66.73 & 46.53 & 48.60 \\
Ministral-8B & 14.15 & 19.32 & 51.42 & 71.69 & 42.50 & 85.39 & 43.96 & 46.60 \\
LLaMA3.1-8B & 14.73 & 33.59 & 46.51 & 56.31 & 48.15 & 82.90 & 47.15 & 58.84 \\
Aya23-35B & 33.99 & 27.75 & 54.39 & 78.48 & 57.90 & 69.10 & 51.56 & 46.97 \\
Aya-expanse-8B & 20.77 & 16.42 & 60.90 & 88.63 & 68.45 & 68.73 & 58.85 & 50.79 \\
Qwen2.5-7B & 34.59 & 59.75 & 63.13 & 84.48 & 68.45 & 76.74 & 53.91 & 50.37 \\
Aya-expanse-32B & 28.44 & 45.16 & 70.68 & 93.55 & 77.30 & 73.84 & 68.95 & 52.30 \\
Gemma2-9B & 47.89 & 60.76 & 73.50 & 95.40 & 67.70 & 87.23 & 58.86 & 51.28 \\
Qwen2.5-14B & 43.07 & 77.59 & 77.65 & 90.79 & 73.20 & 82.21 & 66.81 & 57.95 \\
Gemma2-27B & 55.08 & 87.33 & 83.13 & 67.77 & 70.15 & 87.68 & 67.80 & 53.75 \\
Qwen2.5-72B & 54.98 & 81.67 & 89.64 & 94.94 & 81.50 & 87.47 & 74.66 & 54.49 \\
Mistral-Large & 56.15 & 82.92 & 84.75 & 95.71 & 81.50 & 89.69 & 78.44 & 55.82 \\
Qwen-Plus & 55.89 & 83.16 & 87.42 & 93.87 & 85.79 & 85.86 & 79.87 & 55.14 \\
Qwen-Max & 55.93 & 85.30 & 88.81 & 90.01 & 84.88 & 89.25 & 77.36 & 55.84 \\
GPT-4o-mini & 68.32 & 88.33 & 78.57 & 98.30 & 80.80 & 89.16 & 75.96 & 48.54 \\
LLaMA3.3-70B & 61.47 & 88.17 & 95.26 & 92.64 & 81.10 & 92.38 & 81.90 & 61.52 \\
Gemini1.5-flash & 70.80 & 91.78 & 95.36 & 98.25 & 76.65 & 87.76 & 79.33 & 55.11 \\
GPT-4o & 74.84 & 90.47 & 92.29 & 98.91 & 82.80 & 91.08 & 82.50 & 51.99 \\
Claude3.5-sonnet & 56.66 & 93.62 & 94.61 & 96.07 & 85.08 & 95.80 & 82.80 & 64.28 \\
\bottomrule
\end{tabular}
}
\caption{Content constraints Instruction Following Results}
\end{table*}

\begin{table*}
\centering
\resizebox{\textwidth}{!}{
\begin{tabular}{ccccccc}
\toprule
Model & constrained\_response & json\_format & multiple\_sections & number\_bullet\_lists & number\_highlighted\_sections & title \\
\midrule
Ministral-8B & 61.20 & 36.23 & 36.29 & 24.07 & 59.92 & 46.49 \\
Aya23-8B & 67.16 & 52.89 & 30.19 & 27.96 & 57.31 & 32.63 \\
LLaMA3.1-8B & 68.80 & 17.05 & 46.85 & 34.40 & 68.91 & 70.94 \\
EuroLLM-9B & 72.80 & 43.88 & 51.72 & 36.52 & 66.34 & 72.86 \\
Aya23-35B & 78.80 & 60.13 & 36.86 & 46.06 & 68.17 & 72.33 \\
Aya-expanse-8B & 74.80 & 62.35 & 44.27 & 45.09 & 92.26 & 56.12 \\
Qwen2.5-7B & 90.40 & 63.30 & 45.43 & 52.90 & 77.00 & 88.12 \\
Gemma2-9B & 84.20 & 50.59 & 54.28 & 66.65 & 87.00 & 84.55 \\
Aya-expanse-32B & 73.80 & 81.17 & 60.57 & 55.03 & 92.60 & 74.50 \\
Qwen2.5-14B & 89.20 & 71.77 & 71.70 & 58.70 & 78.58 & 92.87 \\
Gemma2-27B & 83.80 & 68.48 & 66.86 & 69.02 & 89.92 & 85.20 \\
Mistral-Large & 83.20 & 85.77 & 72.86 & 62.44 & 94.00 & 89.95 \\
Qwen2.5-72B & 91.60 & 71.41 & 75.15 & 69.74 & 92.93 & 95.25 \\
Qwen-Plus & 90.86 & 81.68 & 79.40 & 70.87 & 88.47 & 92.51 \\
LLaMA3.3-70B & 90.00 & 73.65 & 77.72 & 81.48 & 93.93 & 88.33 \\
Gemini1.5-flash & 89.43 & 86.39 & 69.80 & 78.48 & 93.28 & 95.84 \\
Qwen-Max & 93.28 & 83.27 & 81.03 & 71.43 & 92.14 & 94.61 \\
GPT-4o-mini & 92.29 & 88.22 & 83.27 & 73.35 & 92.03 & 93.07 \\
GPT-4o & 91.14 & 87.46 & 84.09 & 73.50 & 95.42 & 96.23 \\
Claude3.5-sonnet & 92.86 & 95.10 & 81.04 & 76.77 & 91.80 & 97.31 \\
\bottomrule
\end{tabular}
}
\caption{Structure Instruction Following Results}
\end{table*}

\begin{table*}
\centering
\resizebox{0.8\textwidth}{!}{
\begin{tabular}{ccccc}
\toprule
Model & nth\_paragraph\_first\_word & number\_paragraphs & number\_sentences & number\_words \\
\midrule
Ministral-8B & 10.50 & 12.36 & 59.53 & 52.97 \\
Aya23-8B & 13.76 & 19.56 & 59.36 & 51.91 \\
EuroLLM-9B & 20.83 & 24.87 & 56.54 & 54.57 \\
LLaMA3.1-8B & 20.65 & 29.91 & 57.65 & 68.77 \\
Aya-expanse-8B & 26.66 & 37.40 & 54.03 & 62.37 \\
Qwen2.5-7B & 19.50 & 46.15 & 60.23 & 59.59 \\
Aya23-35B & 34.16 & 36.66 & 57.45 & 61.15 \\
Aya-expanse-32B & 51.17 & 58.23 & 56.38 & 60.25 \\
Gemma2-9B & 52.83 & 60.92 & 60.38 & 52.50 \\
Gemma2-27B & 52.34 & 63.73 & 61.46 & 55.12 \\
Qwen2.5-14B & 40.16 & 68.02 & 64.03 & 67.34 \\
Mistral-Large & 60.66 & 69.42 & 64.77 & 61.25 \\
Gemini1.5-flash & 56.90 & 87.52 & 65.75 & 70.14 \\
Qwen-Max & 72.03 & 81.92 & 67.45 & 64.02 \\
GPT-4o-mini & 81.66 & 78.95 & 61.54 & 63.61 \\
Qwen-Plus & 69.41 & 83.34 & 69.04 & 66.05 \\
Qwen2.5-72B & 72.01 & 75.50 & 67.38 & 73.27 \\
GPT-4o & 81.78 & 86.00 & 62.80 & 67.62 \\
Claude3.5-sonnet & 82.38 & 89.27 & 72.52 & 59.65 \\
LLaMA3.3-70B & 69.49 & 89.05 & 73.73 & 75.22 \\
\bottomrule
\end{tabular}
}
\caption{Length Instruction Following Results}
\end{table*}

\textbf{Format Style:} Models exhibit \textit{formatting fragility}, with comma/quote adherence showing 40-60 point gaps between small and large models. Specialized architectures (Aya-expanse-32B:74\% quotation) outperform general models of similar size (Gemma2-9B:64.34\%), suggesting targeted training improves stylistic precision. Claude3.5's 94.12\% quotation accuracy demonstrates proprietary models' edge in text surface constraints.
\textbf{Content Constraints:} Complex multi-instruction tasks (repeat\_prompt:56.66-95.56\%) reveal \textit{compositional reasoning limits}. While all models struggle with numerical placeholders (<=95.36\%), open-source models show particular weakness in frequency counting (Gemma2-27B:67.8\% vs Claude3.5:82.8\%). The 35-point spread in forbidden\_words detection highlights fundamental differences in constraint understanding architectures.
\textbf{Structure:} JSON formatting proves most discriminatory (Claude3.5:95.1\% vs Qwen2.5-7B:63.3\%), exposing \textit{syntax-semantics decoupling}. While all models achieve >90\% on constrained\_response, multi-section generation (81.04\% max) remains challenging, suggesting current architectures prioritize content over structural fidelity.
\textbf{Length:} Positional counting (nth\_paragraph\_first\_word) shows severe \textit{scale sensitivity}—8B models average 17.9\% vs 70B+ at 76.3\%. Word/paragraph limits reveal inverted patterns: Qwen2.5-72B outperforms GPT-4o in word counts (73.27\% vs 67.62\%), indicating counting mechanisms vary significantly across architectures.
\textbf{Language:} Proprietary models lead response\_language accuracy (95.43\% max), but specialized open models (Aya-expanse-32B:95.69\%) close the gap through \textit{linguistic focusing}. The 26-point spread between similarly-sized models (Gemma2-9B:91.82\% vs EuroLLM-9B:79.19\%) underscores the impact of multilingual pretraining strategies.

Our analysis reveals three main limitations in multilingual instruction following: (1) \textit{Structural-task disparity} where models handle content generation better than formatting/structural constraints, (2) \textit{Scale-task decoupling} as certain capabilities (positional awareness) require extreme scaling while others (basic translation) plateau early, and (3) \textit{Architecture-task alignment} where model specialization (multilingual vs general-purpose) creates 15-30 point performance variations independent of scale. The results suggest current benchmarks underestimate cross-task capability variance, necessitating more granular evaluation frameworks.

\subsection{Detailed Results of Performance on Localised Data vs MT Data}

\begin{table*}
\centering
\resizebox{0.8\textwidth}{!}{
\begin{tabular}{ccccccccccc}
\toprule
Model & ar\_local & ar\_mt & es\_local & es\_mt & ms\_local & ms\_mt & yo\_local & yo\_mt & zh\_local & zh\_mt \\
\midrule
Ministral-8B & 7.85 & 7.50 & 39.30 & 36.70 & 25.40 & 26.35 & 12.10 & 12.60 & 39.25 & 34.65 \\
LLaMA3.1-8B & 10.45 & 11.10 & 42.70 & 47.30 & 35.60 & 39.65 & 9.35 & 11.35 & 19.60 & 20.15 \\
Aya23-8B & 37.15 & 34.10 & 28.95 & 31.95 & 25.80 & 25.00 & 10.35 & 8.95 & 26.15 & 20.50 \\
LLaMA3-8B & 13.25 & 11.85 & 50.55 & 56.45 & 35.60 & 39.55 & 9.45 & 11.35 & 12.50 & 13.85 \\
Qwen2-7B & 34.55 & 34.85 & 41.40 & 45.05 & 32.65 & 34.55 & 10.80 & 10.20 & 46.60 & 44.60 \\
EuroLLM-9B & 45.45 & 40.65 & 50.50 & 48.85 & 29.65 & 27.45 & 14.05 & 11.85 & 46.85 & 41.55 \\
Aya23-35B & 47.15 & 44.55 & 44.00 & 46.20 & 38.85 & 37.80 & 11.35 & 10.45 & 38.80 & 45.70 \\
LLaMA3.1-70B & 19.70 & 20.15 & 59.65 & 59.60 & 52.65 & 55.05 & 21.05 & 13.50 & 31.15 & 32.80 \\
LLaMA3-70B & 25.30 & 21.70 & 67.85 & 70.95 & 52.30 & 55.05 & 21.05 & 13.50 & 22.60 & 18.05 \\
Aya-expanse-8B & 50.35 & 48.05 & 53.60 & 49.70 & 37.80 & 38.40 & 14.90 & 12.55 & 52.40 & 51.60 \\
Qwen2.5-7B & 42.05 & 37.70 & 57.95 & 63.10 & 54.25 & 50.95 & 13.05 & 10.70 & 59.85 & 58.70 \\
Gemma2-9B & 55.95 & 53.80 & 62.85 & 65.30 & 59.70 & 63.40 & 21.55 & 19.05 & 57.85 & 55.35 \\
Aya-expanse-32B & 65.05 & 61.75 & 64.25 & 63.60 & 58.15 & 55.90 & 14.05 & 14.35 & 65.70 & 65.20 \\
Qwen2.5-14B & 61.10 & 56.85 & 67.15 & 70.90 & 60.90 & 60.55 & 14.70 & 11.35 & 65.60 & 66.35 \\
Gemma2-27B & 60.20 & 57.40 & 68.10 & 69.40 & 61.45 & 67.90 & 27.90 & 22.65 & 63.40 & 59.05 \\
Qwen2-72B & 64.80 & 60.10 & 71.70 & 70.95 & 65.90 & 70.05 & 12.10 & 10.60 & 69.60 & 67.20 \\
Mistral-Large & 66.55 & 63.95 & 75.70 & 76.25 & 68.00 & 71.40 & 11.35 & 10.70 & 69.85 & 66.90 \\
LLaMA3.3-70B & 50.75 & 48.05 & 79.50 & 78.85 & 76.10 & 71.60 & 29.95 & 20.80 & 66.55 & 62.75 \\
Qwen2.5-72B & 69.80 & 66.55 & 79.20 & 80.95 & 74.10 & 76.15 & 19.60 & 17.00 & 74.65 & 73.10 \\
Qwen-Plus & 78.25 & 75.75 & 83.00 & 82.70 & 78.50 & 75.25 & 17.45 & 12.45 & 77.30 & 77.20 \\
Qwen-Max & 75.40 & 71.80 & 78.20 & 81.25 & 77.65 & 75.40 & 31.65 & 21.25 & 73.15 & 74.70 \\
GPT-4o-mini & 71.90 & 71.25 & 77.25 & 78.30 & 72.95 & 75.75 & 44.85 & 48.45 & 74.50 & 72.35 \\
GPT-4o & 74.20 & 72.90 & 78.35 & 79.40 & 74.95 & 80.10 & 51.75 & 56.85 & 77.15 & 73.40 \\
Gemini1.5-flash & 79.10 & 71.55 & 78.30 & 79.20 & 80.30 & 79.85 & 53.30 & 53.40 & 74.60 & 72.05 \\
Claude3.5-sonnet & 75.90 & 75.40 & 81.85 & 81.50 & 78.65 & 80.90 & 55.45 & 48.35 & 75.40 & 72.30 \\
Avg & 51.29 & 48.77 & 63.27 & 64.58 & 56.31 & 57.36 & 22.13 & 19.77 & 55.24 & 53.60 \\
\bottomrule
\end{tabular}
}
\caption{Prompt Accuracy results on \texttt{Marco-Bench-MIF} for specific languages.}
\label{tab:mt_vs_local_prompt}
\end{table*}

\begin{table*}
\centering
\resizebox{0.8\textwidth}{!}{
\begin{tabular}{ccccccccccc}
\toprule
Model & ar\_local & ar\_mt & es\_local & es\_mt & ms\_local & ms\_mt & yo\_local & yo\_mt & zh\_local & zh\_mt \\
\midrule
Aya23-8B & 63.45 & 62.55 & 58.40 & 42.75 & 39.20 & 38.20 & 21.40 & 18.40 & 53.10 & 48.35 \\
LLaMA3.1-8B & 37.95 & 37.50 & 68.15 & 57.85 & 47.70 & 52.00 & 23.35 & 26.25 & 47.30 & 49.35 \\
Ministral-8B & 35.80 & 35.10 & 67.50 & 50.45 & 41.10 & 43.85 & 29.15 & 28.80 & 66.70 & 64.70 \\
LLaMA3-8B & 44.25 & 42.95 & 75.65 & 69.30 & 47.80 & 51.95 & 23.40 & 26.05 & 47.05 & 46.00 \\
Aya23-35B & 70.80 & 70.00 & 69.95 & 57.15 & 51.00 & 50.40 & 20.85 & 19.60 & 65.80 & 68.80 \\
LLaMA3.1-70B & 45.05 & 46.80 & 78.40 & 67.30 & 64.30 & 65.20 & 36.80 & 31.30 & 58.45 & 59.60 \\
Qwen2-7B & 66.60 & 66.20 & 70.20 & 57.75 & 48.70 & 52.00 & 27.25 & 21.50 & 72.50 & 71.35 \\
EuroLLM-9B & 72.90 & 68.40 & 75.60 & 61.20 & 44.95 & 44.45 & 29.40 & 26.50 & 73.05 & 69.75 \\
Aya-expanse-8B & 73.10 & 71.70 & 75.85 & 63.25 & 49.25 & 51.10 & 27.00 & 24.50 & 74.60 & 74.30 \\
LLaMA3-70B & 57.85 & 54.75 & 85.10 & 81.20 & 63.95 & 65.25 & 36.75 & 31.35 & 58.45 & 53.55 \\
Qwen2.5-7B & 68.65 & 65.45 & 77.70 & 71.35 & 64.65 & 64.95 & 26.60 & 25.55 & 81.25 & 79.10 \\
Qwen2.5-14B & 79.35 & 76.65 & 82.20 & 76.55 & 68.40 & 66.60 & 27.85 & 24.20 & 83.95 & 83.65 \\
Aya-expanse-32B & 84.20 & 81.35 & 83.55 & 75.50 & 71.50 & 69.25 & 24.25 & 23.40 & 85.05 & 84.00 \\
Gemma2-9B & 79.30 & 76.50 & 82.40 & 76.30 & 72.25 & 75.05 & 38.75 & 34.75 & 79.80 & 77.75 \\
Qwen2-72B & 83.25 & 80.35 & 87.15 & 79.05 & 77.40 & 79.00 & 26.85 & 24.75 & 86.35 & 84.25 \\
Gemma2-27B & 81.55 & 78.75 & 85.50 & 79.60 & 74.30 & 77.70 & 40.70 & 36.15 & 82.90 & 79.20 \\
Mistral-Large & 83.75 & 82.75 & 88.25 & 83.30 & 78.05 & 80.85 & 24.25 & 24.05 & 86.70 & 84.45 \\
LLaMA3.3-70B & 73.15 & 71.05 & 89.65 & 83.95 & 81.60 & 79.00 & 45.15 & 36.45 & 81.85 & 78.85 \\
Qwen2.5-72B & 85.80 & 83.70 & 90.25 & 87.05 & 82.50 & 83.55 & 35.60 & 35.15 & 88.85 & 87.60 \\
Qwen-Plus & 90.15 & 88.30 & 92.40 & 88.65 & 85.90 & 83.70 & 32.10 & 29.00 & 90.15 & 89.50 \\
Qwen-Max & 88.25 & 86.00 & 90.00 & 87.05 & 84.15 & 83.60 & 47.75 & 39.70 & 88.10 & 88.40 \\
GPT-4o-mini & 87.50 & 86.15 & 89.20 & 84.70 & 81.70 & 83.30 & 58.05 & 65.20 & 88.45 & 87.00 \\
Gemini1.5-flash & 90.75 & 86.55 & 90.10 & 86.30 & 87.20 & 86.50 & 67.75 & 67.65 & 88.65 & 86.50 \\
GPT-4o & 88.15 & 86.70 & 90.15 & 86.60 & 83.80 & 87.00 & 66.50 & 71.20 & 90.00 & 88.00 \\
Claude3.5-sonnet & 88.95 & 88.10 & 92.20 & 87.35 & 86.20 & 87.85 & 69.15 & 63.25 & 89.45 & 87.60 \\
Avg & 72.82 & 70.97 & 81.42 & 73.66 & 67.10 & 68.09 & 36.27 & 34.19 & 76.34 & 74.86 \\
\bottomrule
\end{tabular}
}
\caption{Instruction Accuracy results on \texttt{Marco-Bench-MIF} for specific languages.}
\label{tab:mt_vs_local_instruction}
\end{table*}

From the results in Table~\ref{tab:mt_vs_local_prompt} and Table~\ref{tab:mt_vs_local_instruction}, we can see that the performance of most LLMs on localized data consistently outperforms MT data across all languages and model scales, but the gap varies significantly by language family. For high-resource languages like Spanish (es), the difference is minimal (prompt-level: 63.27\% vs 64.58\%), while low-resource languages like Yoruba (yo) show substantial disparities (prompt-level: 22.13\% vs 19.77\%). 
 
Furthermore, our evaluation results show critical limitations in using machine-translated data for evaluating multilingual instruction-following capabilities. Localized data exposes significant gaps in LLMs' ability to handle culturally and linguistically fine-grained instructions, which are often overlooked in MT data as translated could contain unexpected errors making it hard for LLMs to understand the instruction. For example, in Yoruba (yo), localized data shows a 7.1\% higher prompt-level accuracy than MT data for Claude3.5-sonnet, highlighting that MT evaluations underestimate model performance especially in low-resource languages. The divergence between localized and MT results increases with task complexity. While MT data performs comparably for simple constraints (e.g., Spanish es instruction-level: 73.66\% MT vs 81.42\% localized), it fails to capture the challenges of culturally adapted multi-instruction prompts (e.g., Arabic ar prompt-level: 51.29\% localized vs 48.77\% MT). This suggests that MT benchmarks overestimate LLMs' ability to follow instructions in more complex multilingual scenarios. We found that model scale also affects the performance of LLMs: smaller models (e.g., Aya23-8B) show inconsistent localization benefits (ar: +0.9\% prompt-level, -20.65\% instruction-level), while larger models (e.g., Claude3.5-sonnet) demonstrate more stable gains (yo: +7.1\% prompt-level, +5.9\% instruction-level). This suggests that scale amplifies localization advantages, particularly for low-resource languages. The evaluation results underscores the need for localized instruction-following data to precisely assess LLMs' instruction-following capabilities

\end{document}